\documentclass[conference]{IEEEtran}
\IEEEoverridecommandlockouts
\usepackage{bm}
\newtheorem{definition}{Definition}[section]
\usepackage{multirow}

\newcommand{\mymethod}{aggregated anomaly detection with normalizing flows}
\newcommand{\mymethodAC}{GRADINGS}
\usepackage{cite}
\usepackage{amsmath,amssymb,amsfonts}
\usepackage{graphicx}
\usepackage{subfigure}
\usepackage{textcomp}
\usepackage{booktabs}
\usepackage{xcolor}
\def\BibTeX{{\rm B\kern-.05em{\sc i\kern-.025em b}\kern-.08em
    T\kern-.1667em\lower.7ex\hbox{E}\kern-.125emX}}
    
\usepackage{algorithm}
\usepackage{algcompatible}

\usepackage{tabularx}
\usepackage{booktabs}
\usepackage[version=4]{mhchem}
\newcommand{\bftab}{\fontseries{b}\selectfont}
\newcommand{\rot}[1]{\multirow{9}{*}{\rotatebox[origin=c]{90}{#1}}}
\begin{document}

\title{Anomaly Detection in Trajectory Data\\with Normalizing Flows
}

\makeatletter
\newcommand{\linebreakand}{%
  \end{@IEEEauthorhalign}
  \hfill\mbox{}\par
  \mbox{}\hfill\begin{@IEEEauthorhalign}
}
\makeatother

\author{\IEEEauthorblockN{Madson L. D. Dias}
\IEEEauthorblockA{\textit{Department of Computer Science} \\
\textit{Federal University of Cear{\'a}}\\
Fortaleza, Brazil \\
madsonddias@gmail.com}
\and
\IEEEauthorblockN{C{\'e}sar Lincoln C. Mattos}
\IEEEauthorblockA{\textit{Department of Computer Science} \\
\textit{Federal University of Cear{\'a}}\\
Fortaleza, Brazil \\
cesarlincoln@dc.ufc.br}
\and
\IEEEauthorblockN{Ticiana L. C. da Silva}
\IEEEauthorblockA{\textit{Virtual UFC Institute} \\
\textit{Federal University of Cear{\'a}}\\
Fortaleza, Brazil \\
ticianalc@insightlab.ufc.br}
\linebreakand
\IEEEauthorblockN{Jos{\'e} Ant{\^o}nio F. de Mac{\^e}do}
\IEEEauthorblockA{\textit{Department of Computer Science} \\
\textit{Federal University of Cear{\'a}}\\
Fortaleza, Brazil \\
jose.macedo@insightlab.ufc.br}
\and
\IEEEauthorblockN{Wellington C. P. Silva}
\IEEEauthorblockA{\textit{National Department of Public Security} \\
Federal District, Brazil \\
wellington.wcps@gmail.com}
}

\maketitle

\begin{abstract}
The task of detecting anomalous data patterns is as important in practical applications as challenging. In the context of spatial data, recognition of unexpected trajectories brings additional difficulties, such as high dimensionality and varying pattern lengths. We aim to tackle such a problem from a probability density estimation point of view, since it provides an unsupervised procedure to identify out of distribution samples. More specifically, we pursue an approach based on normalizing flows, a recent framework that enables complex density estimation from data with neural networks. Our proposal computes exact model likelihood values, an important feature of normalizing flows, for each segment of the trajectory. Then, we aggregate the segments' likelihoods into a single coherent trajectory anomaly score. Such a strategy enables handling possibly large sequences with different lengths. We evaluate our methodology, named \mymethod ~(\mymethodAC), using real world trajectory data and compare it with more traditional anomaly detection techniques. The promising results obtained in the performed computational experiments indicate the feasibility of the \mymethodAC, specially the variant that considers autoregressive normalizing flows.
\end{abstract}


\section{Introduction}

The wide availability of spatial data acquisition devices, from specialized remote sensors to standard GPS equipped smartphones, has resulted in the creation of several location-based applications. Although the object to be localized can vary (a vehicle, an animal, a person, etc.), in general, such trajectory data can be understood as a series of ordered points that characterize the object motion \cite{zheng2015trajectory}.

In the context of trajectory data, anomaly detection is a task critical to monitor spatial events and enable recognition of unexpected behaviors \cite{meng2019overview}. One could define an anomaly, or outlier\footnote{In this work we use the terms \textit{outlier} and \textit{anomaly} interchangeably.}, as a data point which significantly differs from the overall observed data \cite{aggarwal2015outlier}. It is worth emphasizing that, as opposed to single point standard regression, in such a scenario a data example is a full trajectory or at least a segment of it.

Meng \textit{et al.} \cite{meng2019overview} propose a traditional anomaly detection taxonomy that includes methods based on classification, clustering, distance, density and statistics. We pursue the latter, which consists in a model-based procedure that aims to explain the available data, mostly within a probabilistic density estimation framework. Anomalies are then detected by measuring how much the model fits a given new data point. Such an approach does not require labeled data, as it is a form of unsupervised learning. However, probabilistic density estimation approaches for trajectory anomaly detection usually consider simple distributions, such as a multivariate Gaussian \cite{hazel2000multivariate}. Even if a more flexible Gaussian mixture model (GMM) is used, such as in \cite{basharat2008learning,li2016anomaly}, it is not straightforward to determine the number of components in the mixture.

In this work we tackle the task of trajectory data analysis by pursuing an approach based on normalizing flows (NFs,~\cite{rezende2015variational}), a general framework for estimating complex probabilistic densities. In summary, a NF transforms an initial simple density by a sequence of invertible transformations to better explain the observed data. Following recent works, such as \cite{rezende2015variational}, each transformation, i.e. a flow, is parametrized by (possibly deep) neural networks. One of the main advantages of such flow-based approach is the available exact model log-likelihood, which is used as an objective function to jointly optimize all the model parameters. Furthermore, we can compute exact log-likelihood values for new data points, which we will then apply as a coherent anomaly score.

NF-based anomaly detection approaches have been recently proposed \cite{yamaguchi2019adaflow,iwata2019supervised}. In contrast to those works, we aim to evaluate NF models with trajectory data, which is inherently sequential. Thus, we include in our evaluations the so-called masked autoregressive flow (MAF), an autoregressive flow framework that directly models the conditional distributions of the input variables \cite{papamakarios2017masked}.

Trajectory data can be high dimensional due to the presence of several measured points within a single trajectory. Besides, distinct data examples can have different lengths, which cannot be straightforwardly compared. We propose a methodology that tackles both issues by considering fixed-size segments of the available trajectories. A NF generative model is than used to estimate the probability density of such segments. It is expected that segments which belong to trajectories considered normal correspond to higher model likelihoods than segments that belong to trajectories considered anomalous. In the test step, a single anomaly score for a new trajectory is computed from its segments using an aggregation function. Moreover, since we choose a trajectory data representation that incorporates timestamps, the time domain is considered in the modeling. We name our approach \mymethod ~(\mymethodAC). We emphasize that, to the best of our knowledge, our work is the first evaluation of NF-based models in the task of anomaly detection in trajectory data.

We evaluate the proposed \mymethodAC ~approach using real-world trajectories available in the Microsoft GeoLife data set \cite{zheng2009mining,zheng2008,zheng2010}. The obtained experimental results indicate the feasibility of our solution. Our framework, specially the variant that considers the MAF model, achieves better anomaly detection results in comparison to standard techniques, such as the GMM and the local outlier factor (LOF,~\cite{breunig2000lof}) method.

\section{Problem statement and data representation}\label{sec:problem}

The problem of anomaly detection can be vaguely described as the task of finding data patterns that differ from what is expected and is considered normal \cite{agrawal2015survey}. In this work, such unexpected patterns are related to trajectories sufficiently different from the previously seen data, which is assumed to be mostly normal. We consider that trajectories can differ in terms of spatial segments that comprise them and/or the time period they occur.

As follows we establish the adopted data representation and the main theoretical aspects of the unsupervised anomaly detection task.

\subsection{Trajectory representation}


Broadly speaking, a trajectory consists of a sequence of GPS points (i.e., latitude, longitude and timestamp) generated by a moving object on a monitoring system. Below we formally define it.

\begin{definition}[Trajectory] A trajectory $\mathbf{T}_m \in \mathcal{T}$ with size $L_m$ is defined as a finite ordered sequence
\begin{equation}
\mathbf{T}_m \triangleq \left( \bm{q}_1^{(m)}, \bm{q}_2^{(m)},\cdots,\bm{q}_l^{(m)},\cdots, \bm{q}_{L_m}^{(m)} \right),    
\end{equation}
where $\bm{q}_l^{(m)} = \left( q_{l,1}^{(m)}, q_{l,2}^{(m)}, q_{l,3}^{(m)} \right)$ is a \emph{location point}, such that $q_{l,1}^{(m)}, q_{l,2}^{(m)}, q_{l,3}^{(m)}$ are respectively the $l$-th latitude, $l$-th longitude, and $l$-th timestamp of the trajectory $\mathbf{T}_m$. Furthermore, we have $q_{l_0,3} < q_{l_1,3}$, for all $l_0 < l_1$, which ensures a temporal ordered sequence of points.
\end{definition}



\subsection{Problem statement}

Given a set of trajectories, the goal of a trajectory anomaly detection model is to find trajectories that are significantly different from the majority, considered to be normal. In other words, let~$\mathcal{T} = \{\bm{\mathrm{T}}_n\}_{n=1}^N$ be a set of trajectories of moving objects in a GPS monitoring system. The task of trajectory anomaly detection is to create a model from the available trajectories to evaluate the anomaly degree of any given trajectory~$\bm{\mathrm{T}}$.

In this work, we follow an unsupervised anomaly detection procedure, which does not require the trajectories to be previously labeled as normal or anomalous. We detail such an approach as follows.

\subsection{Unsupervised anomaly detection}

Unsupervised anomaly detection approaches (or anomaly detection over noisy data~\cite{eskin2000}) makes two assumptions over the data. The first one is that the dataset contains a large number of normal elements and relatively few anomalies. The second assumption is that the abnormal data is generated by a different probability distribution~\cite{leung2005}.

After the training step, the determination if a sample $\bm{x}$ is normal or abnormal can be made by a decision system $H$ as follows:
\begin{equation}
    \label{eq:anomaly_decision}
    H(\bm{x}, \phi) =
    \begin{cases}
    ~0 \text{ (normal) }& \text{ if }~~~ A(\bm{x}) < \phi,\\
    ~1 \text{ (abnormal) }& \text{ if }~~~ A(\bm{x}) \geq \phi,
    \end{cases}
\end{equation}
where $\phi$ is a predefined threshold and $A(\cdot)$ is an \emph{anomaly score}. The threshold $\phi$ is a value that separates abnormal from normal data samples. In the context of supervised and semi-supervised anomaly detection, this value is usually chosen by using a validation set that contains known anomalous samples~\cite{schmidt2019,neema2019}. After that, metrics such as accuracy and $F_1$-score are computed to judge the quality of the models.

Alternatively, and more common to the unsupervised learning setup, we can judge the model quality without the choice of a single value for the threshold $\phi$. This can be done by finding the receiver operating characteristic (ROC) curve, which indicates the relation between the false positive rate and the true positive rate as the threshold is changed. A practical metric to summarize the information provided by the ROC curve is the area under the curve (AUC or AUROC)~\cite{ling2003}.

\section{Classical anomaly detection techniques}

Anomaly detection algorithms can be classified in several groups based on distance, probability, reconstruction, and information theory~\cite{pimentel2014}. In this section, we describe two of most know techniques used in anomaly detection problems.

\subsection{Anomaly detection using the LOF algorithm}


Local outlier factor~(LOF,~\cite{breunig2000lof}) is an unsupervised distance-based anomaly detection algorithm. The anomaly score in LOF is computed by comparing the \emph{local density} of a sample to the surrounding neighborhood. The local density is inversely correlated with the average distance from the point to its neighborhood. 

Let $\mathcal{X}$ be a set of data points. The set of $K$-nearest neighbors of $\bm{x} \in \mathcal{X}$ is denoted by $\mathcal{N}_{\bm{x}}^{k}$ and defined as $\mathcal{N}_{\bm{x}}^{K} \triangleq k\mathrm{NN}\left(K, \bm{x}, \mathcal{X} \setminus \{\bm{x}\}\right)$, where $k\mathrm{NN}\left(\cdot, \cdot, \cdot \right)$ is the result of a $K$-nearest neighbor query~\cite{roussopoulos1995,Papadias2009}. 

Then, we define the $K$-distance neighborhood~$\mathrm{KD}(\bm{x})$ of a sample $\bm{x} \in \mathcal{X}$ as $\mathrm{KD}(\bm{x}) \triangleq \max_{\bm{u} \in \mathcal{N}_{\bm{x}}^{K}} \|\bm{x} - \bm{u} \|,$ where $\|\cdot\|$ is the Euclidean distance.

We use the above to define the reachability distance $\mathrm{RD}(\bm{x}, \bm{u})$ of $\bm{x}$ with respect to another sample $\bm{u} \in \mathcal{X}$ as $\mathrm{RD}(\bm{x}, \bm{u}) \triangleq \max  \left \{\mathrm{KD}(\bm{x}), \|\bm{x} - \bm{u} \| \right \}$.

The local reachability density of $\bm{x}$ with respect to $\bm{u}$ is then denoted by $\mathrm{LRD}(\bm{x}, \bm{u})$ and defined as
\begin{equation*}
    \mathrm{LRD}(\bm{x}) \triangleq \dfrac{K}{\sum_{\bm{u} \in \mathcal{N}_{\bm{x}}^{K}} \mathrm{RD}(\bm{x}, \bm{u})}.
\end{equation*}

Using all the previous definitions, the LOF anomaly score can be finally formalized as the average ratio of local reachability densities with respect to $\bm{x}$ and its $K$-neighborhood:
\begin{equation}
    A(\bm{x}) = \frac{1}{K} \sum_{\bm{u} \in \mathcal{N}_{\bm{x}}^{K}} \frac{\mathrm{LRD}(\bm{x})}{\mathrm{LRD}(\bm{u})}.
\end{equation}
Note that the above score measures the local density deviation of a given data point with respect to its neighbors.

\subsection{Gaussian mixture model for anomaly detection}

One way of computing an anomaly score $A$ in Eq. \eqref{eq:anomaly_decision} is to use a probability density estimator. This approach first trains the density estimator $p(\cdot)$ and then uses the negative log-likelihood of each testing data as an anomaly score, i.e.,
\begin{equation}
    A(\bm{x}) = - \ln p(\bm{x}).
\end{equation}

In such a context, the Gaussian mixture model (GMM) is a common choice. A GMM uses a linear combination of Gaussian density functions to approximate an unknown probability distribution. The parameters of each the component are usually adjusted using the Expectation Maximization~(EM,~\cite{dempster1977}) algorithm.

Consider a data set $\mathcal{D} = \{\bm{x}_n\}_{n=1}^N$, where $\bm{x}_n \in \mathbb{R}^D$. We assume that the points from $\mathcal{D}$ are generated in an i.i.d. fashion from an underlying density $p(\bm{x})$. Furthermore, suppose that $p(\bm{x})$ is defined as a finite mixture model with $K$ components:
\begin{equation}
\label{eq_gmm}
p(\bm{x}) = \sum_{k=1}^K \pi_k \mathcal{N}(\bm{x} | \bm{\mu}_k, \mathbf{\Sigma}_k)
\end{equation}
where $\mathcal{N}(\bm{x} | \bm{\mu}_k, \mathbf{\Sigma}_k)$ is a multivariate Gaussian density with mean vector $\bm{\mu}_k$ and covariance matrix $\mathbf{\Sigma}_k$; $\{\pi_k\}_{k=1}^K$ are the mixture weights, which are restricted to be non-negative and sum up to 1, i.e., $\sum_{k=1}^K \pi_k = 1$. The mixture weights represent the probability that a randomly selected data point $\bm{x}$ was generated by the component $k$. After the optimization of the GMM parameters via the EM algorithm, Eq. \eqref{eq_gmm} can be directly applied as an anomaly score for new data points.

It is worth noting that flow-based generative models constitute a flexible alternative to density estimation with standard techniques such as the GMM. In the next section we detail the flow-based models used in this work.

\section{Probability density estimation via normalizing flows}\label{sec:flows}

NF models are powerful tools for estimating complicated probability densities~\cite{zhisheng2019,kingma2018glow}. Two merits of these models are the exact inference and log-likelihood evaluation~\cite{kingma2018glow}. The latter is specially valuable in the context of anomaly detection.

Let $\bm{x} \in \mathbb{R}^D$ be a random vector with unknown distribution $p(\bm{x})$. In the most general flow-based model, the generative process is defined as~\cite{rezende2015}
\begin{align}
    \bm{h} &\sim p(\bm{h}),\\
    \bm{x} &= g(\bm{h}),
\end{align}
where $\bm{h}$ is a latent (unobserved) variable and $p(\bm{h})$ is a simple and known distribution, e.g., a multivariate Gaussian. The function $g(\cdot)$, called \emph{bijective}, is an invertible function such that $g^{-1}(\bm{x}) = f(\bm{x}) = \bm{h}$. If the transformation $f(\cdot)$ is considered to be a composition of $K$ successive mappings and we apply the change of variables rule, the log-likelihood of the random variable $\bm{x}$ can be written as~\cite{rezende2015}
\begin{equation}
    \ln p_K(\bm{z}_K) = \ln p_0 (\bm{z}_0) - \sum_{k=1}^K \ln \left| \det \frac{\partial \bm{z}_{k}}{\partial \bm{z}_{k-1}}  \right|,
\end{equation}
where $\bm{x} \triangleq \bm{z}_K \sim p_K(\bm{z}_K)$, $\bm{h} \triangleq \bm{z}_0 \sim p_0(\bm{z}_0)$ and $\bm{z}_k = f_k(\bm{z}_{k-1}), \forall k=1,2,\dots,K$. 

The usual training criterion of flow-based generative models is simply the negative log-likelyhood over the training set $\mathcal{X}$:
\begin{equation}
    \label{eq_nf_loss}
    L(\mathcal{X}) = -\frac{1}{|\mathcal{X}|} \sum_{\bm{x} \in \mathcal{X}} \ln p(\bm{x}).
\end{equation}

We summarize the evaluated NF models as follows.





\subsection{Real NVP}

Real-valued non-volume preserving (Real-NVP,~\cite{dinh2017density}) is a type of NF that uses a bijection called \emph{coupling layer} that transforms only some input dimensions via functions that depend on the untransformed dimensions. If $1:d$ denotes the sequential indexes of the $d$ untransformed dimensions, the components of the layer output $\bm{y}$ are given by
\begin{align}
    \bm{y}_{1:d}   &= \bm{x}_{1:d}, \\
    \bm{y}_{d+1:D} &= \bm{x}_{d+1:D} \odot \exp \left( \sigma(\bm{x}_{1:d}) \right) + \mu(\bm{x}_{1:d}),
\end{align}
where $\sigma, \mu: \mathbb{R}^d \to \mathbb{R}^{D-d}$ respectively represent scale and translation functions parametrized by neural networks, and $\odot$ is the element-wise product operator. The elements in each flow are permuted to different orders, allowing all of the inputs to have a chance to be altered.


The Jacobian matrix of the above described transformations can be calculated using
\begin{equation}\label{eq:rnvp-determinant}
    \frac{\partial \bm{y}}{\partial \bm{x}} = 
        \begin{bmatrix}
            \mathbf{I}_d & \mathbf{0}_{d\times(D-d)} \\
            \frac{\partial \bm{y}_{d+1:D}}{\partial \bm{x}_{1:d}} & \mathrm{diag}\left(\exp\left(\sigma(\bm{x}_{1:d})\right)\right)
\end{bmatrix},
\end{equation}
where $\mathbf{I}_d \in \mathbb{R}^{d \times d}$ is the $d$-order identity matrix, $\mathbf{0}_{d \times (D-d)} \in \mathbb{R}^{d\times(D-d)}$ is a zero matrix and $\mathrm{diag} \left(\exp\left(\sigma(\bm{x}_{1:d})\right)\right) \in \mathbb{R}^{(D-d) \times (D-d)}$ is a diagonal matrix whose elements are equal to the vector $\exp\left(\sigma(\bm{x}_{1:d})\right)$. The Jacobian matrix in Eq. \eqref{eq:rnvp-determinant} is triangular, thus, its determinant is a simple product of the diagonal terms:
\begin{equation}
    \small \det \frac{\partial \bm{y}}{\partial \bm{x}} = \prod_{j=1}^{D-d} \exp\left(\sigma(\bm{x}_{1:d})\right)_j
                      = \exp\left(\sum_{j=1}^{D-d} \sigma(\bm{x}_{1:d})_j\right).
\end{equation}
Since the computation of the Jacobian determinant of the mentioned transformations does not involve calculating the inverse of the functions $\sigma(\cdot) $ and $\mu(\cdot)$, such functions can be arbitrarily complex, usually a deep neural network~\cite{dinh2017density}. All the model parameters (i.e., the networks' weights) are jointly optimized via maximization of the Eq. \eqref{eq_nf_loss} via stochastic gradient descent methods.



\subsection{Masked autoregressive flow (MAF)}

We can decompose any joint density $p(\bm{x})$ of high-dimensional data into a product of one-dimensional conditionals using the chain rule of probabilities: 
\begin{equation}
    \small p(\bm{x}) = \prod_{d=1}^D p(x_d | x_1, x_2, \cdots, x_{d-1}) = \prod_{d=1}^D p(x_d | \bm{x}_{1:d-1}).
\end{equation}

The Masked Autoregressive Flow (MAF, \cite{papamakarios2017masked, germain2015made}) uses the above autoregressive constraint to model the probability density whose conditionals are parameterized as single Gaussians. Thus, the $d$-th conditional probability is given by
\begin{equation}
   \small p(x_d | \bm{x}_{1:d-1}) = \mathcal{N}\left(x_d | \mu_d(\bm{x}_{1:d-1}), \left(\exp \left(\alpha_d(\bm{x}_{1:d-1})\right)\right)^2\right),
\end{equation}
where $\mu_d, \alpha_d: \mathbb{R}^{d-1} \mapsto \mathbb{R}$ are two unconstrained scalar functions that compute the mean and log-standard deviation of the $d$-th conditional given all previous variables. The bijective transformation of MAF generates each $y_d$ conditioned on the past dimensions $\bm{y}_{1:d-1}$, 
\begin{equation}
    y_d = x_d \exp \left( \alpha_d(\bm{y}_{1:d-1}) \right) + \mu_d(\bm{y}_{1:d-1}).
\end{equation}
As a consequence of the autoregressive nature of this transformation, the dimension $d$ of the resulting variable $\bm{y}$ depends only on the $1:d$ dimensions of the input variable $\bm{x}$. Thus, the Jacobian matrix of this transformation is triangular~\cite{kingma2016} and its determinant is equal to the product of its diagonal terms: 
\begin{equation}
    \small \det \frac{\partial \bm{y}}{\partial \bm{x}} = \prod_{d=1}^{D} \exp\left(\alpha_d(\bm{y}_{1:d-1})\right) = \exp\left(\sum_{d=1}^{D} \alpha_d(\bm{y}_{1:d-1})\right).
\end{equation}

As in the RealNVP, the functions $\mu_d(\cdot)$ and $\alpha_d(\cdot)$ can be arbitrarily complex. In the MAF model, these functions are implemented by an efficient feedforward network called Masked Autoencoder for Distribution Estimation (MADE,~\cite{germain2015made}) that takes $\bm{x}$ as input and outputs the means and log-standard deviations for all dimensions in a single network pass.

The nature of MAF transformations allows more flexible generalizations when compared to the RealNVP model. As one can see, if for the first $j \leq d$ dimensions we fix $\mu_j = \alpha_j = 0$ and apply the MAF transformations into the other $j > d$ dimensions, the MAF structure becomes equivalent to the RealNVP. Besides, we can see the coupling layer of the RealNVP as a special case of the MAF transformation~\cite{papamakarios2017masked}.




\section{Proposed Methodology}

We can compute an anomaly score for a sequential data type sample either directly or by first computing scores for local subsections and then aggregating them. These subsections are called pattern fragments, segments, sliding windows, motifs, or n-grams~\cite{gupta2014}. In Definition \ref{def:segment} we present a formal description of these objects.

\begin{definition}[trajectory segment]\label{def:segment}
Given a trajectory $\mathbf{T}_m$ with length $L_m$, the segment~$\mathbf{S}_i$ of $\mathbf{T}_m$ with a user-defined length $W$ is a finite ordered sequence of location points, denoted by
\vspace{-1em}
\begin{equation}
     \mathbf{S}_i^{(m)} \triangleq \left( \bm{q}_{i}^{(m)}, \bm{q}^{(m)}_{i+1}, \cdots, \bm{q}^{(m)}_{i+W} \right).
\end{equation}
where $W \leqslant L_m$ and $1 \leqslant i \leqslant L_m - W + 1$.
\end{definition}

Segment-based techniques usually perform better when compared to direct detection methods~\cite{gupta2014}. Furthermore, they enable handling large sequences with different lengths. As follows we detail our proposal, named \mymethod~(\mymethodAC), which consists in three main steps.

In the first step, \mymethodAC ~transforms the set of trajectories into a set of trajectory segments. Thus, given a set of trajectories $\mathcal{T} = \left\{ \mathbf{T}_m \right\}_{m=1}^M$, the transformed set is defined by 
\vspace{-0.5em}
\begin{equation}
    \mathcal{X} = \bigcup_{m=1}^{M}  \left \{ \bm{x}_{n} = \delta \left( \mathbf{S}_i^{(m)} \right) \middle|_{i=1}^{L_m - W + 1} \right \},
\end{equation}
where $\bm{x}_n \in \mathbb{R}^D$, $1\leqslant n \leqslant N = \sum_{n=1}^N (L_m - W)$, and $\delta(\cdot)$ is a function that flattens a $W \times 3$-segment into a $D$-dimensional row vector, where $D=3W$, i.e.,
$$\small \delta \left(\mathbf{S}_i^{(m)} \right) = \left( q_{i,1}^{(m)}, q_{i,2}^{(m)}, q_{i,3}^{(m)}, \cdots , q_{i+W,1}^{(m)}, q_{i+W,2}^{(m)}, q_{i+W,3}^{(m)}  \right).$$

The second step consists in estimating the distribution $p(\cdot)$ from the available trajectory segments. This step is performed by using one of the NF generative models described in Section~\ref{sec:flows}. At this point, the \mymethodAC~is able to compute the anomaly degree for any trajectory segment, denoted by $\alpha \left(\mathbf{S}_i^{(m)}\right)$:
\vspace{-1em}
\begin{equation}
    \alpha \left(\mathbf{S}_i^{(m)}\right) = - \ln p\left( \delta \left(\mathbf{S}_i^{(m)} \right) \right).
\end{equation}

In the last step, we aggregate the anomaly scores of the segments to compute a single anomaly score for the trajectory. More specifically, given a trajectory $\mathbf{T}_m$, its anomaly score, denoted by $A\left( \mathbf{T}_m \right)$, can be computed using an aggregation function $\varphi$ that combines the anomaly degree of each segment $\mathbf{S}_i^{(n)}$ in the trajectory $\mathbf{T}_m$, i.e.,
\begin{equation}
    \label{eq_aggreg}
    A\left( \mathbf{T}_m \right) = \varphi \left( \left \{ \alpha \left(\mathbf{S}_i^{(m)}\right) \right \}_{i=1}^{L_m - W + 1}  \right).
\end{equation}
Possible choices for the aggregation function $\varphi$ includes the median or the average.



\section{Experiments}

To assess the performance of the proposed methodology, we conduct experiments comparing \mymethodAC~when using either Real NVP or MAF estimators against standard LOF and GMM anomaly detectors with real world data.

\subsection{Data set description}

We consider the version 1.3 of the Microsoft GeoLife data set \cite{zheng2009mining,zheng2008,zheng2010}, comprised of real trajectory data measured from 182 users over a period of five years (from April 2007 to August 2012), which is equivalent to $17621$ trajectories. 
For 73 users, the transportation mode is labeled, such as driving, taking a bus, riding a bike and walking. Each trajectory represents a complete trip from departure to arrival location. 

In our experimental setting, we use a subset of the data that consists of the trajectories located in Beijing, China, made using car (126 trajectories) or bus (365 trajectories). We define two different scenarios. In the first one, called CAR $\times$ BUS, we use the car trajectories as in-distribution data (i.e., as ``normal'' patterns) and the bus trajectories as out-of-distribution data (i.e., as ``anomalies''). In the second one, called BUS $\times$ CAR scenario we switch the roles: the bus trajectories act as in-distribution data and the car trajectories are seen as out-of-distribution samples. For each scenarios we use segments with length correspondent to $10$, $20$, and $30$ location points, accounting a total of $6$ data sets. All of these data sets have $230632$ segments of car trajectories and $850082$ segments of bus trajectories.

The timestamp information of the trajectory data is firstly converted to the hour of the week (e.g. Tuesday, 12:30, is equal to 36.5 if we consider the Monday as the start of the week) and then encoded into two variables using $\left( \sin\left(2 \pi \frac{hour}{168}\right), \cos\left(2 \pi \frac{hour}{168}\right) \right)$. This encoding ensures that similar periodic times are close in the input space, even in different weeks (e.g. Sunday, 23:59 is close to Monday, 00:00).

\subsection{Results and discussion}

We report results for individual segments scores and full trajectories scores. In the latter, we consider both the average and the median as score aggregation functions $\varphi$ (see Eq. \eqref{eq_aggreg}).


We train all the models on the normal data and then apply them to unseen normal samples as well as abnormal data samples. The normal data have been partitioned into two folds, the first one with 80\% of the data for the training, and the other 20\% is grouped with the abnormal data to compute the evaluation metrics.

For the MAF and RealNVP models we use $10$ flows of neural networks as bijective functions, with the MADE structure in the case of the MAF model. Each network has two hidden hidden layers, each one with $32$ neurons. Both models were trained for 300 epochs. A grid search with $5$-fold cross-validation is used to perform the hyper-parameter tuning using the training data for the GMM model. The $K$ value of the LOF algorithm was determined using the heuristic presented in~\cite{breunig2000lof}.

The ROC curves and the correspondent AUROC values are presented in Figs.~\ref{fig:roc-1} and~\ref{fig:roc-2}. In addition, we present in Table~\ref{tab:fpr80} the the false positive rate obtained when we fix a true positive rate of $80\%$, named the FPR80 metric.

In all evaluated pair scenario-variant the NF-based solutions performed better in terms of AUCROC. In most of them, the \mymethodAC ~framework with the MAF model was the best. In terms of FPR80, the MAF also achieved better results in 16 out of 18 evaluations, with the RealNVP being slightly better in the others. It is important to highlight that the use of a segment aggregation strategy considerably increased the performance concerning the AUROC in all experiments. Particularly, models with the median as the aggregation function achieved the best results in terms of AUROC and FPR80.

In terms of the segment length, when using the median as the aggregation function, we can see that the performance is inversely proportional to the size of the segment. On the other hand, using the average as the aggregation function, the performance decreases as the segment size increases. Since the average score is more sensitive to outliers, we hypothesize that the increase of the segment size may cause more outliers to appear in the same pattern. The results that consider only the individual segments do not show any specific behavior with respect to the segment length.

In summary, the obtained results indicate the importance of both main ingredients of the proposed \mymethodAC ~framework: (\textit{i}) the NF-based density estimation; and (\textit{ii}) the aggregation of the individual segments degrees into a single trajectory anomaly score. Furthermore, we have also verified that, in general, the combination of the autoregressive MAF model, the median aggregation function and a larger (e.g. 30) segment length representation offers the best performance.

\begin{table}
\begin{center}

\caption{False positive rates obtained when we fix a true positive rate of $80\%$ (FPR80) for all experimental scenarios.}\label{tab:fpr80}
\begin{tabular}{l l l r r r r }

\toprule

&                        &                & \multicolumn{4}{c}{\emph{Model}}                    \\ \cmidrule{4-7}
\emph{Scenario}        &\emph{Variant}   & \emph{Length}      &        MAF &    RealNVP &        GMM &        LOF  \\ \midrule

\rot{CAR $\times$ BUS} & segment          & $10$           &\bftab 0.423 &       0.643 &      0.698 &      0.719 \\
                       &                  & $20$           &\bftab 0.498 &       0.640 &      0.653 &      0.688 \\
                       &                  & $30$           &\bftab 0.608 &       0.652 &      0.699 &      0.727 \\ \cmidrule{2-7}

                       & average          & $10$           &       0.342 &\bftab 0.335 &      0.376 &      0.465 \\
                       &                  & $20$           &\bftab 0.272 &       0.435 &      0.500 &      0.550 \\
                       &                  & $30$           &\bftab 0.361 &       0.577 &      0.556 &      0.622 \\ \cmidrule{2-7}

                       & median           & $10$           &\bftab 0.245 &       0.375 &      0.308 &      0.481 \\
                       &                  & $20$           &\bftab 0.247 &       0.335 &      0.353 &      0.419 \\
                       &                  & $30$           &\bftab \underline{0.201} &       0.361 &      0.315 &      0.462 \\ \midrule

\rot{BUS $\times$ CAR} & segment          & $10$           &       0.603 &\bftab 0.592 &      0.597 &      0.684 \\
                       &                  & $20$           &\bftab 0.510 &       0.633 &      0.682 &      0.692 \\
                       &                  & $30$           &\bftab 0.489 &       0.517 &      0.631 &      0.689 \\ \cmidrule{2-7}

                       & average          & $10$           &\bftab 0.252 &       0.310 &      0.482 &      0.712 \\
                       &                  & $20$           &\bftab 0.529 &       0.601 &      0.635 &      0.704 \\
                       &                  & $30$           &\bftab 0.311 &       0.555 &      0.622 &      0.732 \\ \cmidrule{2-7}

                       & median           & $10$           &\bftab 0.226 &       0.330 &      0.761 &      0.771 \\
                       &                  & $20$           &\bftab 0.190 &       0.294 &      0.744 &      0.819 \\
                       &                  & $30$           &\bftab \underline{0.055} &       0.328 &      0.564 &      0.747 \\ \bottomrule

\end{tabular}
\vspace{-3em}
\end{center}
\end{table}




\begin{figure*}
     \centering
     \subfigure[]{
     \includegraphics[width=0.3\textwidth]{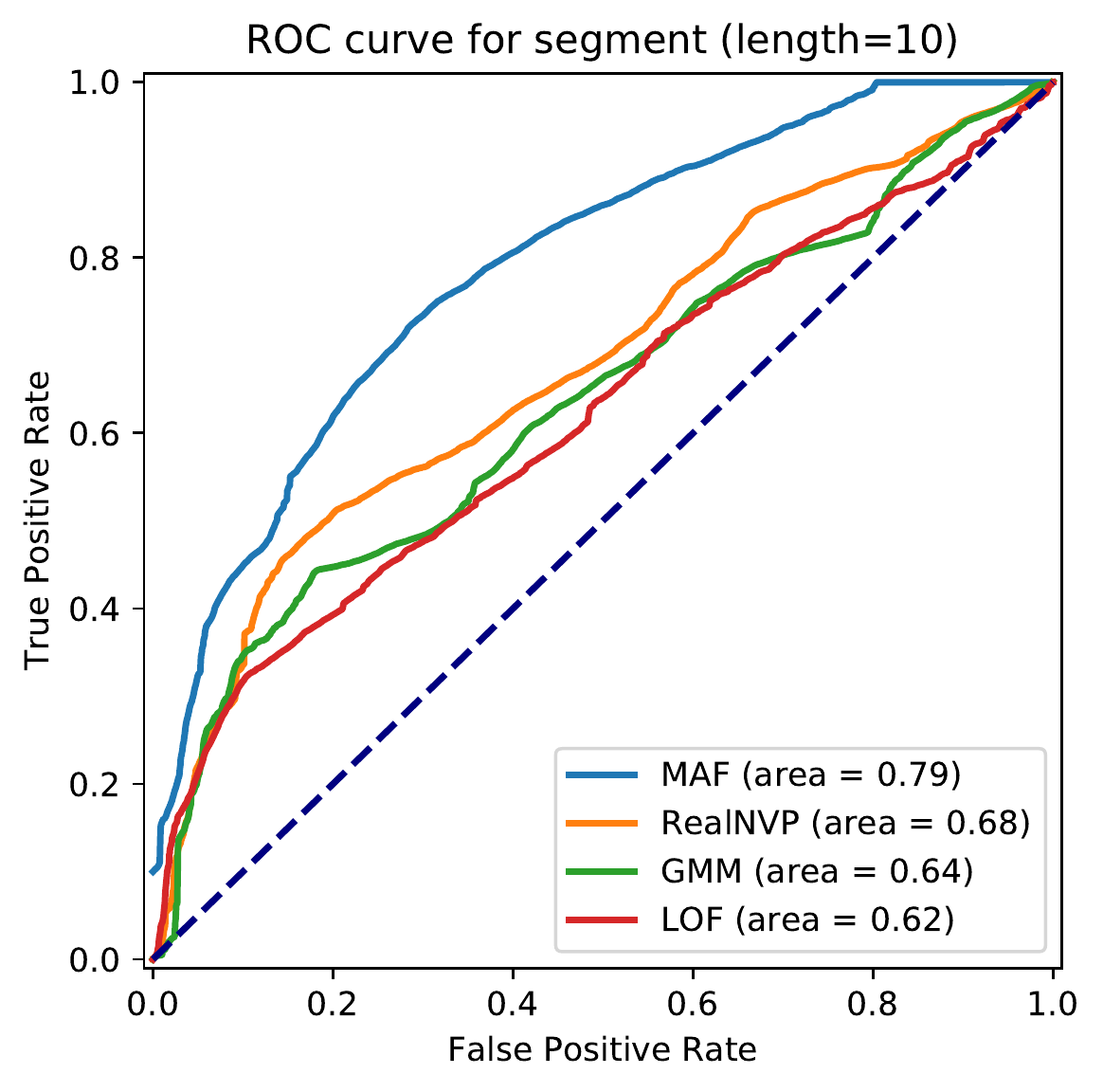}}
     \subfigure[]{\includegraphics[width=0.3\textwidth]{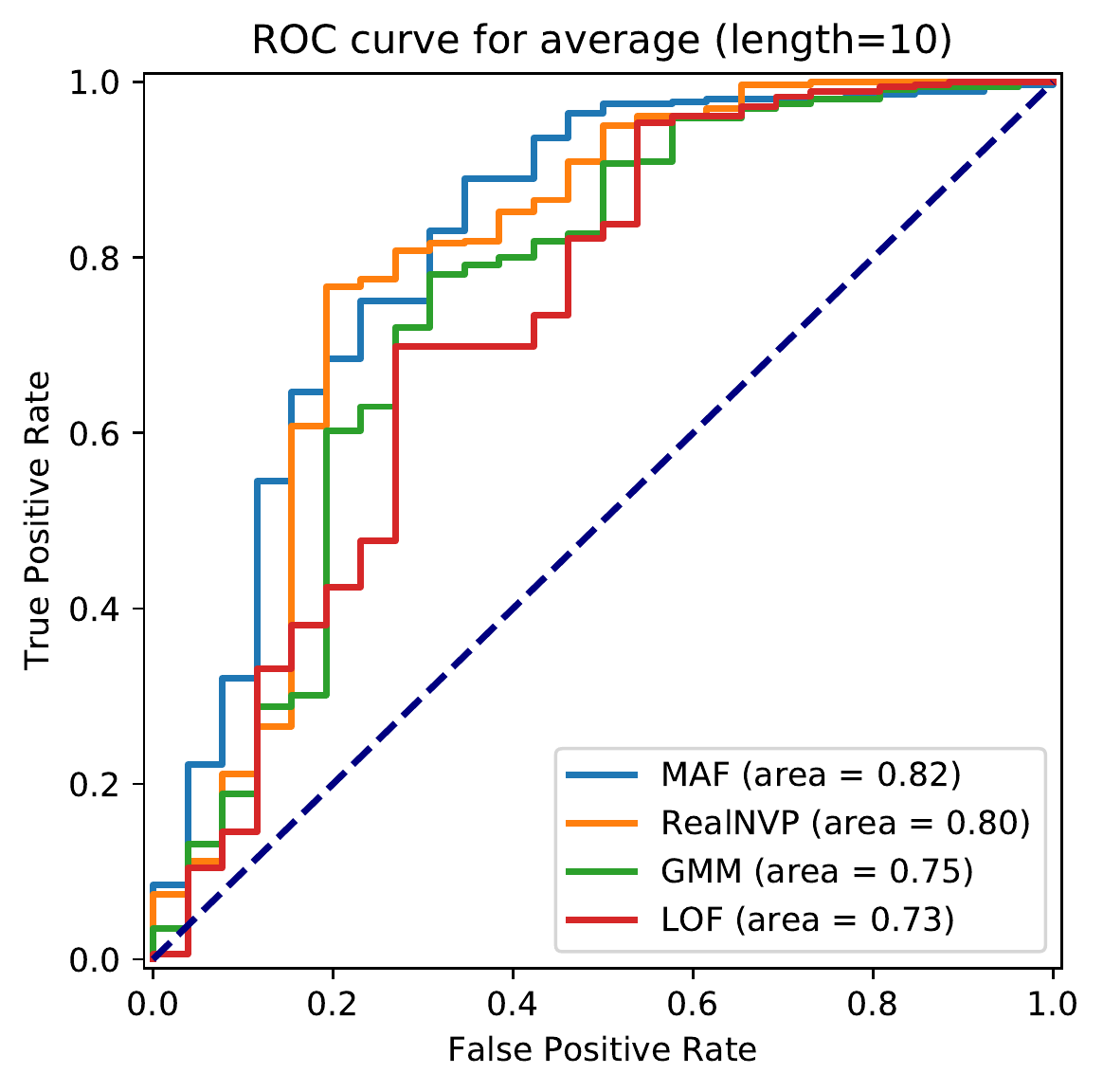}}
     \subfigure[]{\includegraphics[width=0.3\textwidth]{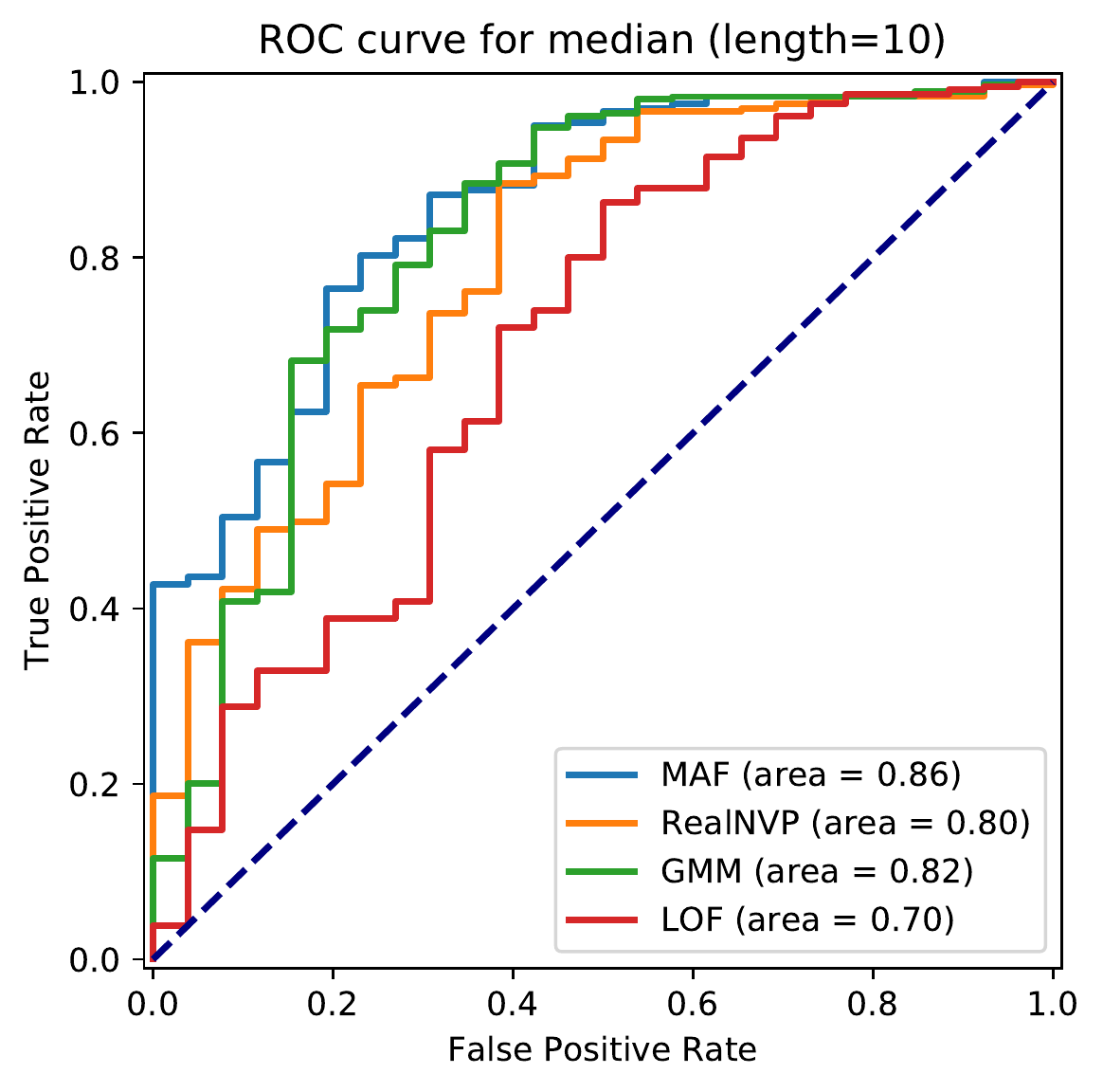}}
     
     \subfigure[]{
     \includegraphics[width=0.3\textwidth]{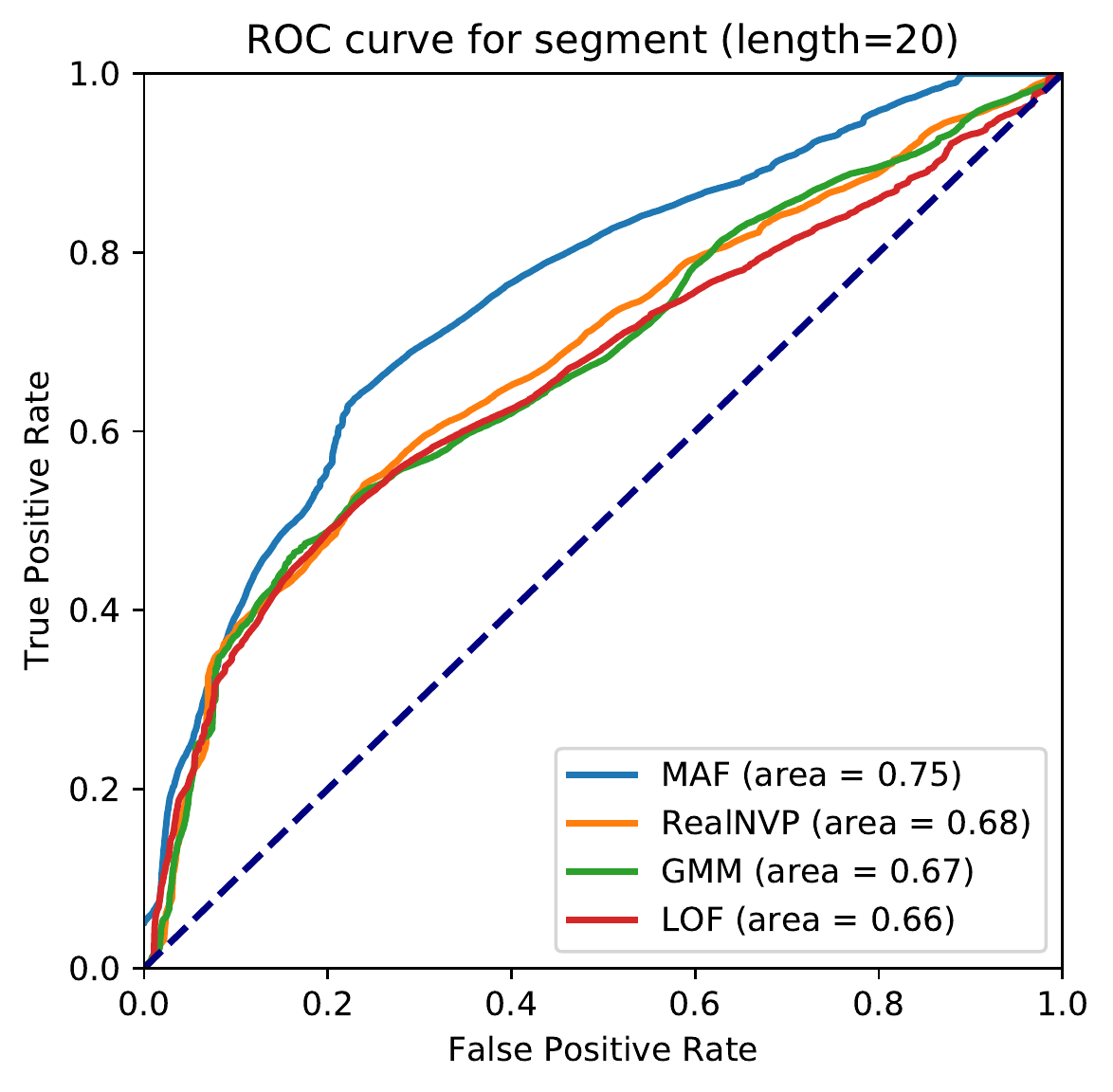}}
     \subfigure[]{\includegraphics[width=0.3\textwidth]{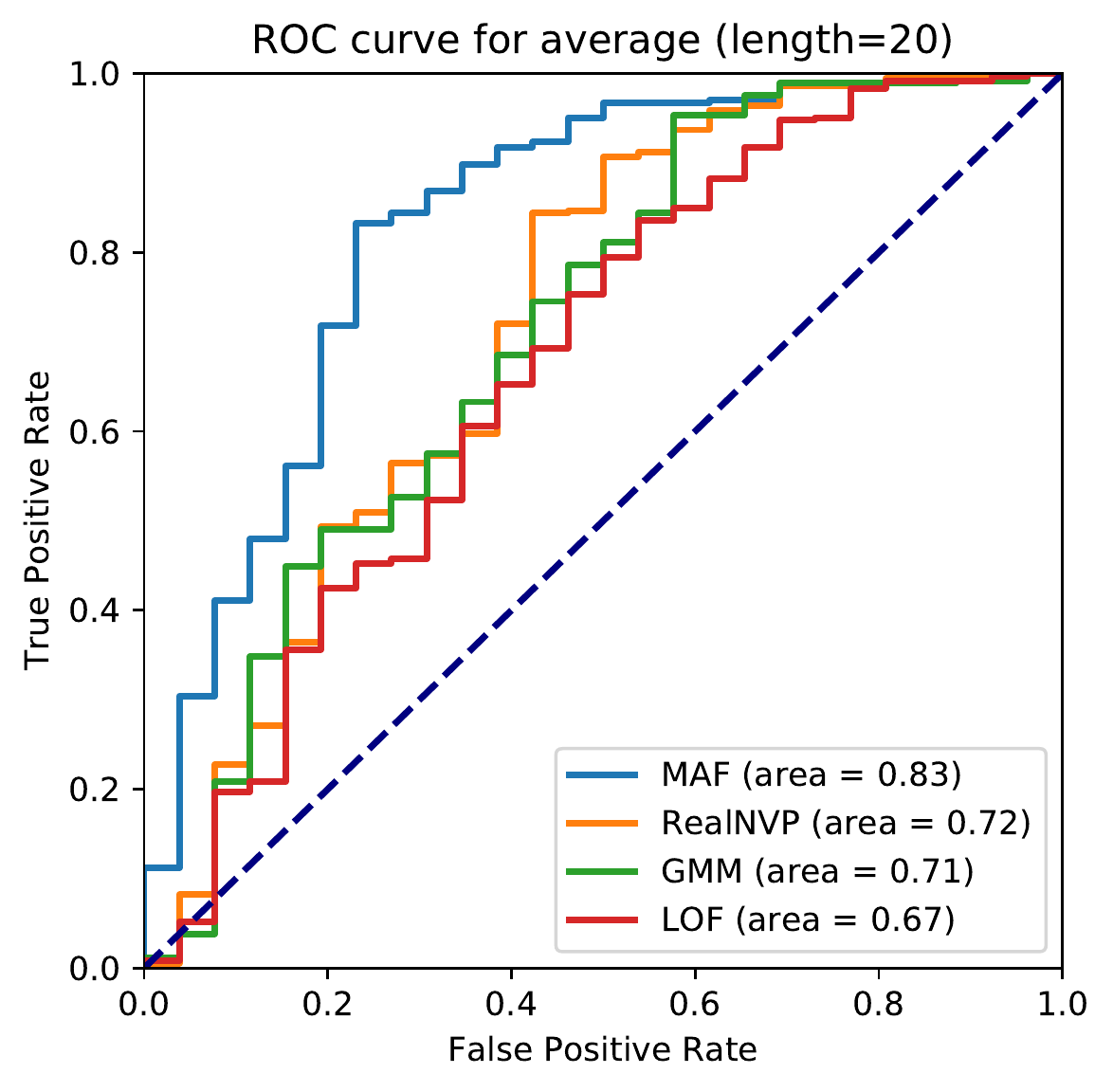}}
     \subfigure[]{\includegraphics[width=0.3\textwidth]{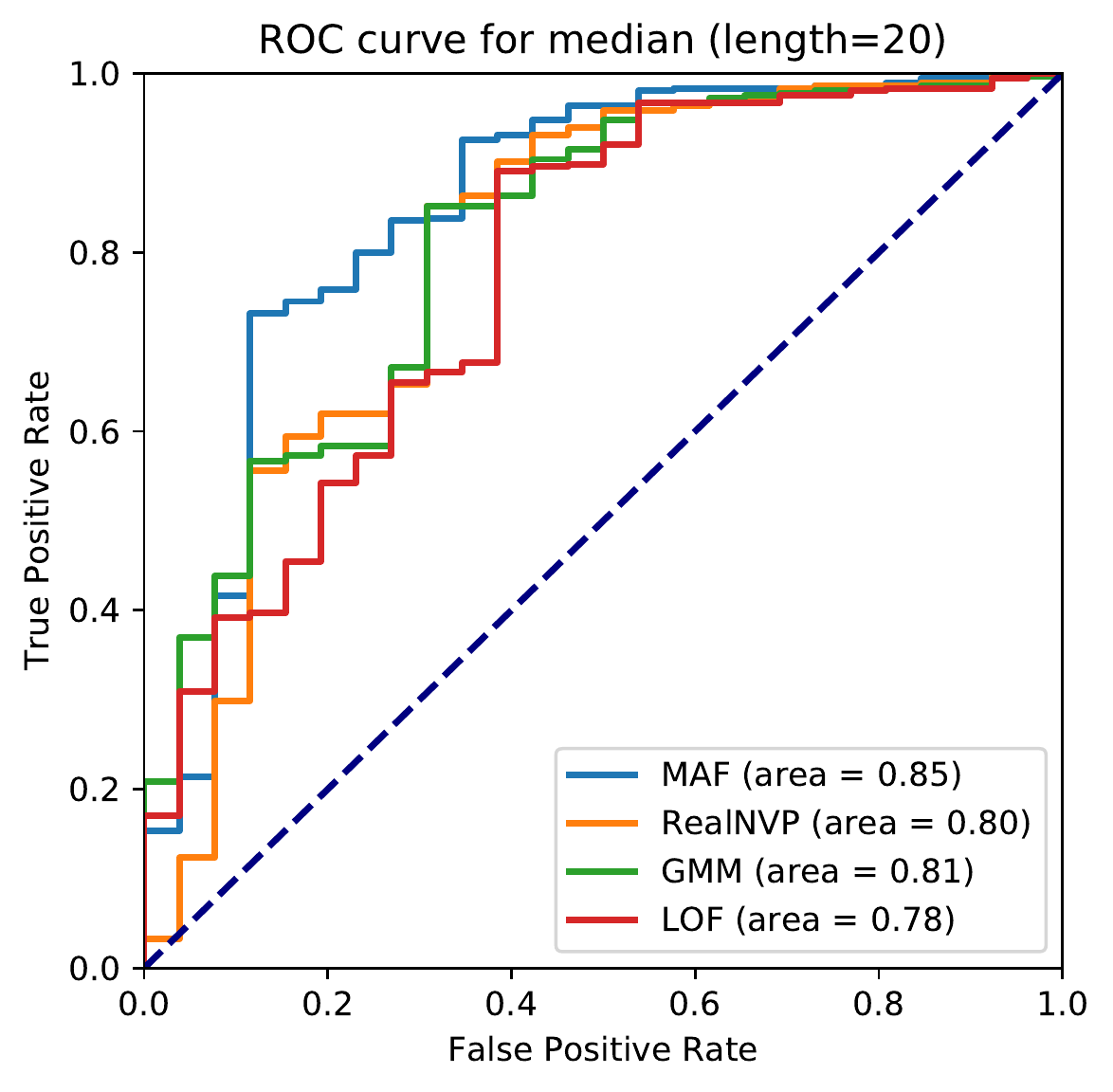}}
     
     \subfigure[]{
     \includegraphics[width=0.3\textwidth]{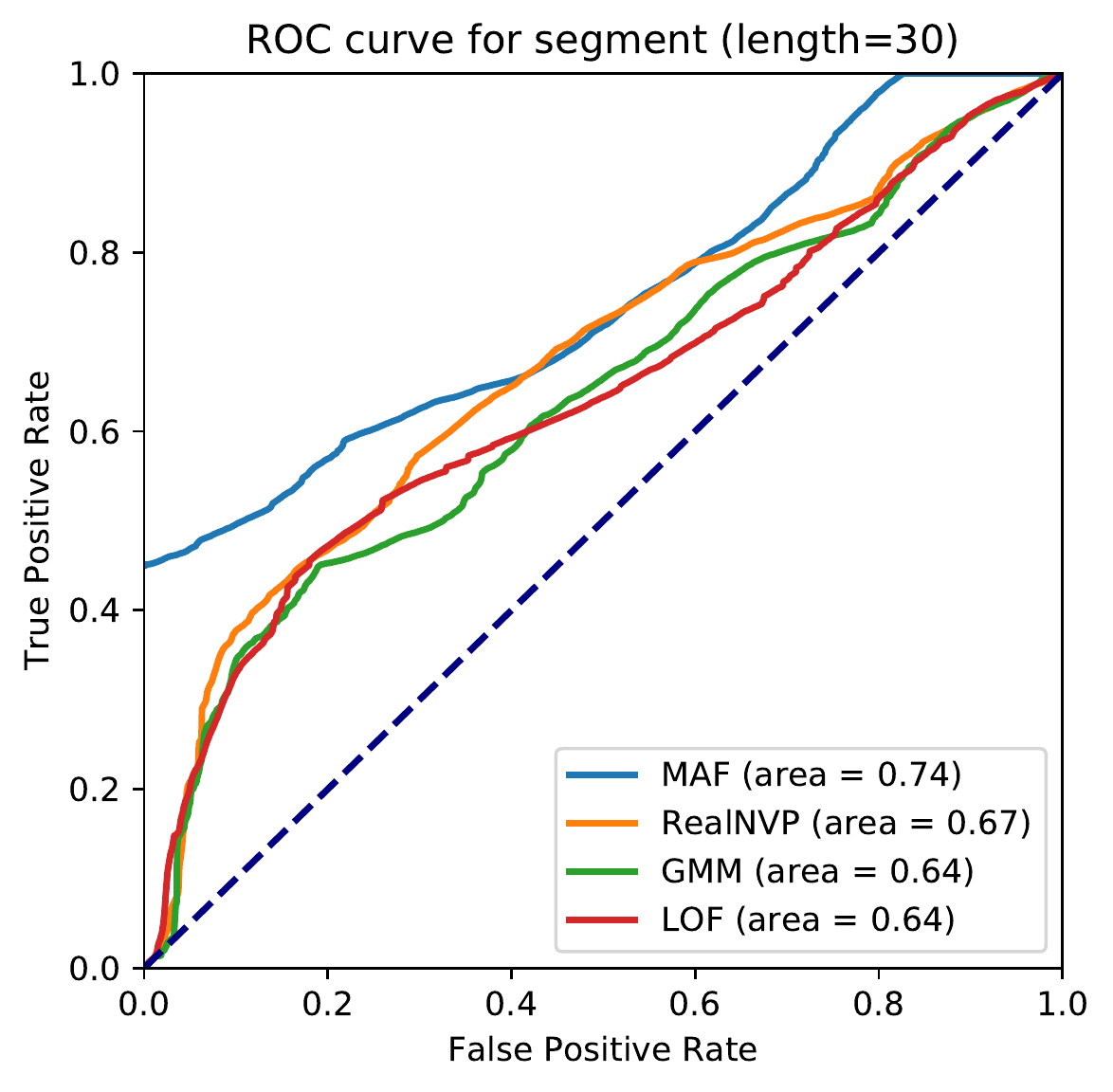}}
     \subfigure[]{\includegraphics[width=0.3\textwidth]{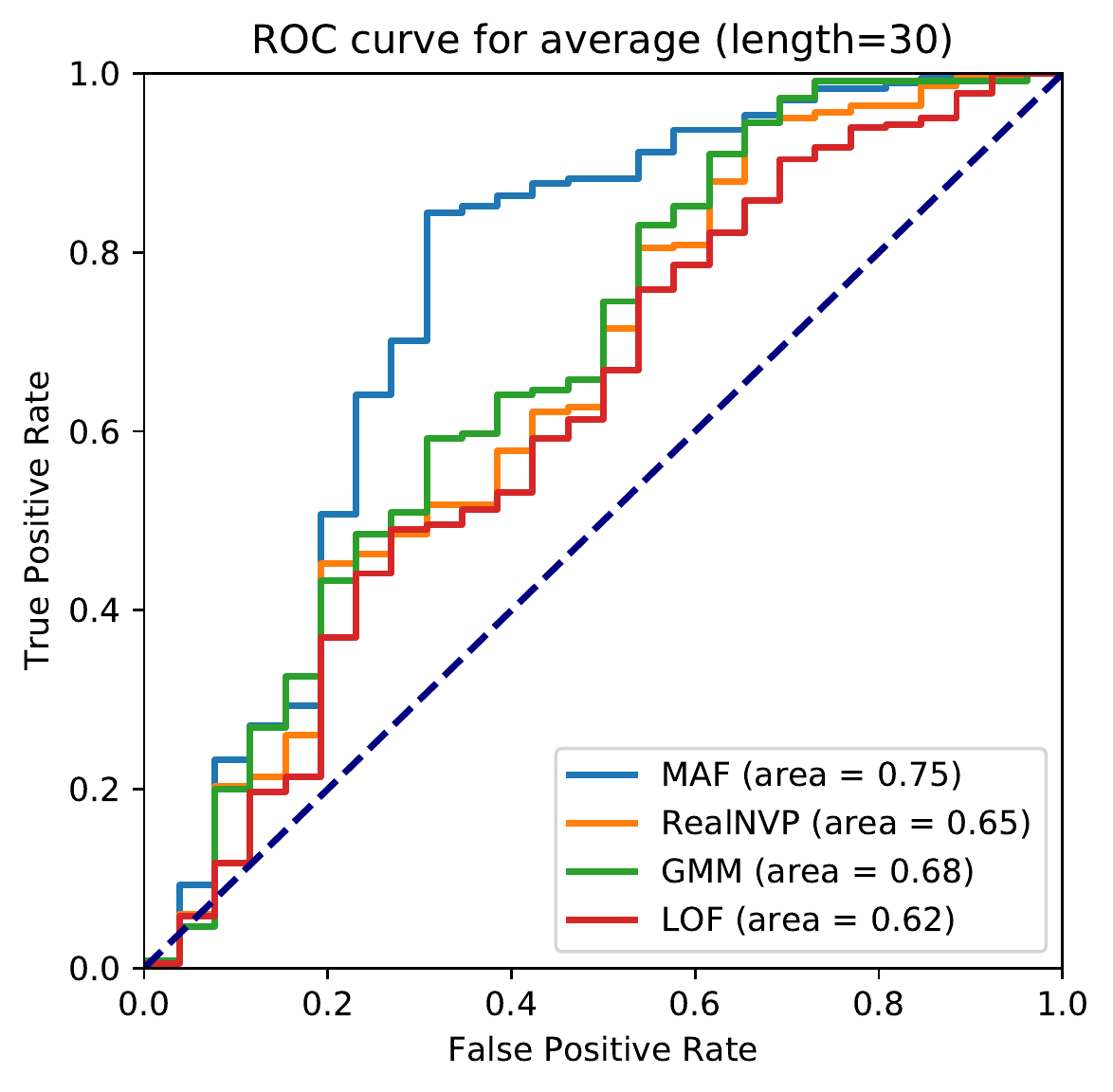}}
     \subfigure[]{\includegraphics[width=0.3\textwidth]{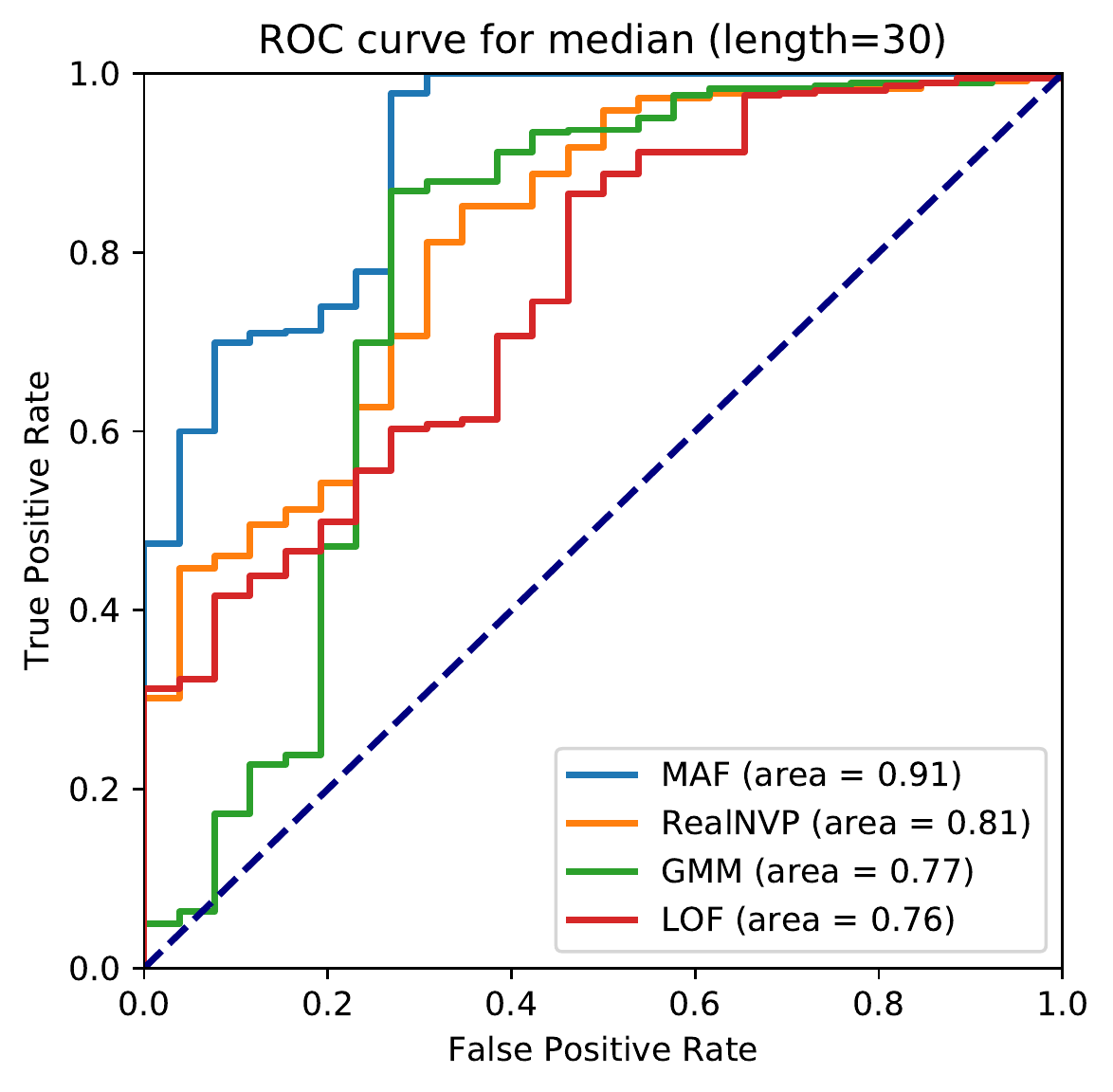}}
        \caption{Anomaly  detection  results  for  CAR $\times$ BUS scenario. ROC curves and respective AUROC values for segments (left column) and for trajectories, using average (middle column), and median (right column). The rows represent the segment lengths -- $10$ (a, b, c), $20$ (d, e, f), and $30$ (g, h, i). The dashed line indicates a completely random detector.}
        \label{fig:roc-1}
\end{figure*}

\begin{figure*}
     \centering
     \subfigure[]{
     \includegraphics[width=0.3\textwidth]{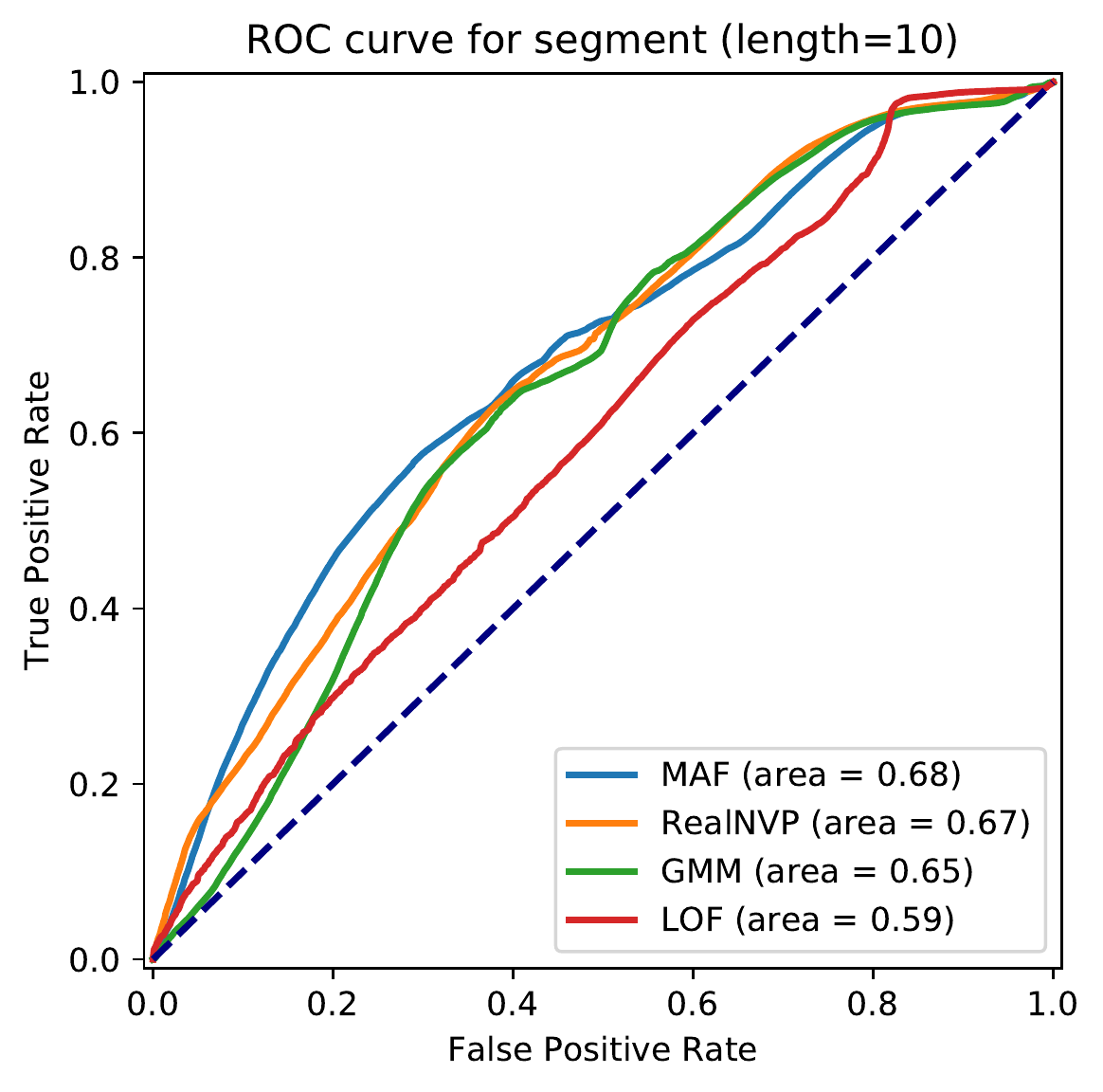}}
     \subfigure[]{\includegraphics[width=0.3\textwidth]{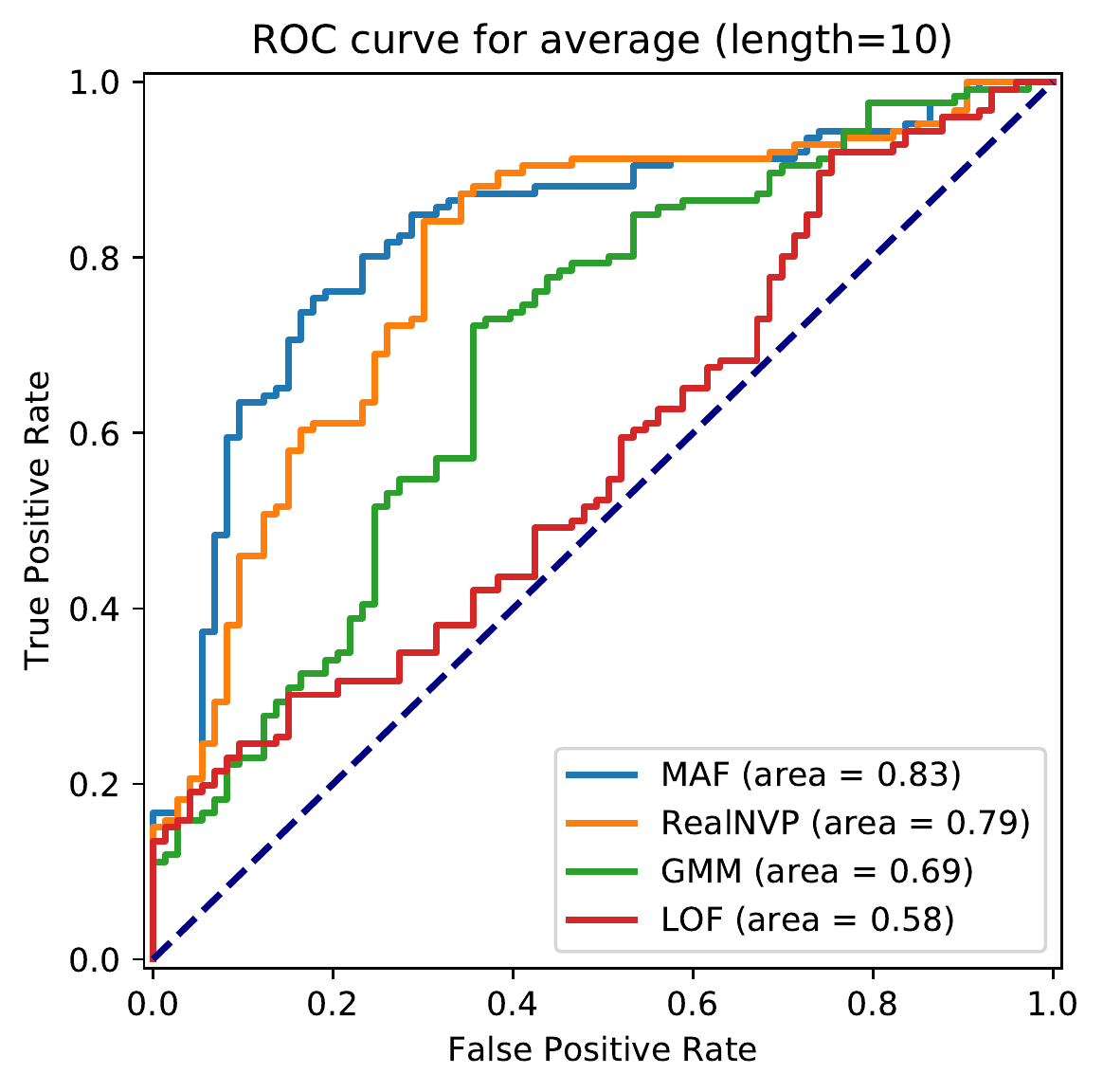}}
     \subfigure[]{\includegraphics[width=0.3\textwidth]{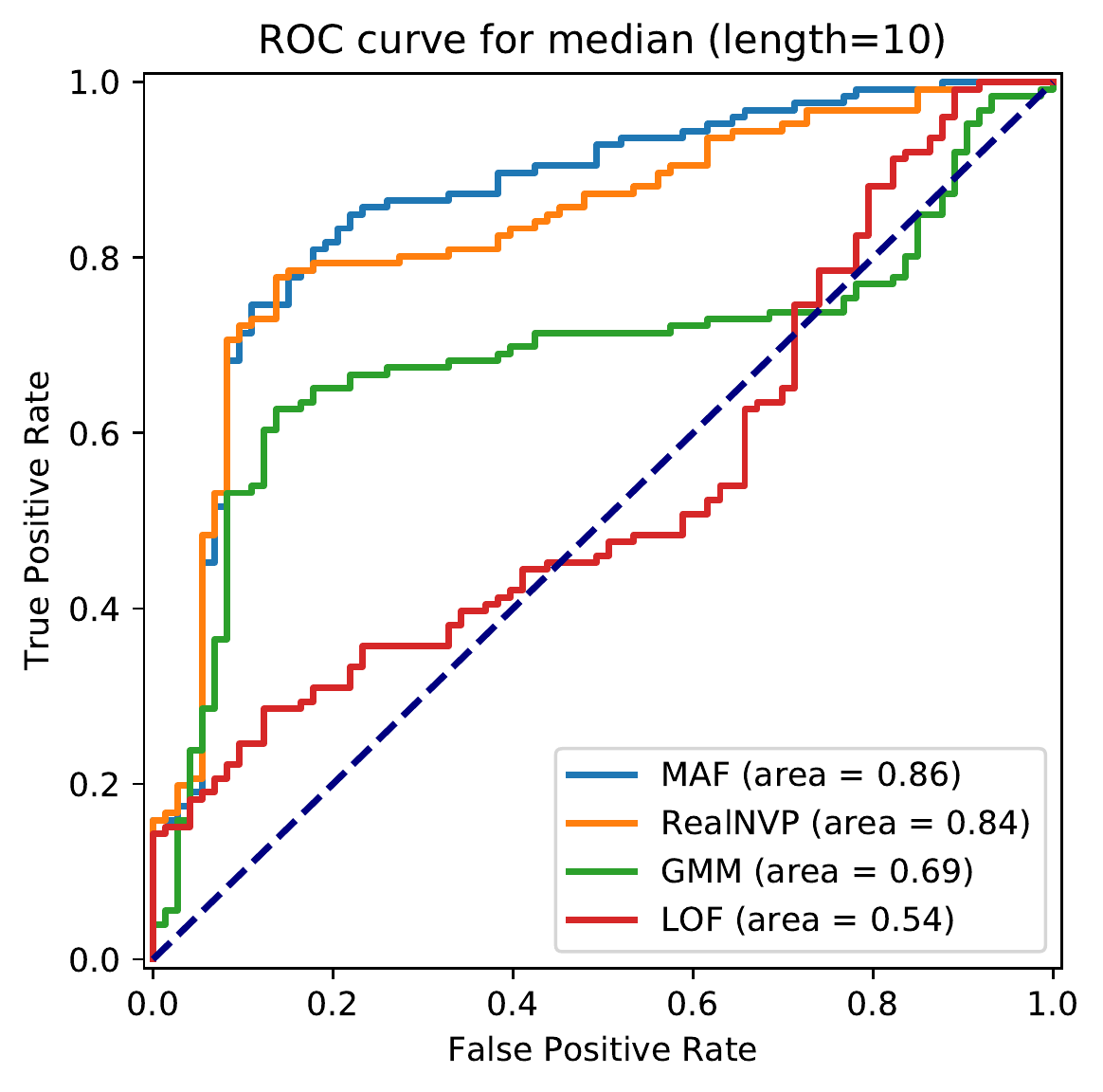}}
     
     \subfigure[]{
     \includegraphics[width=0.3\textwidth]{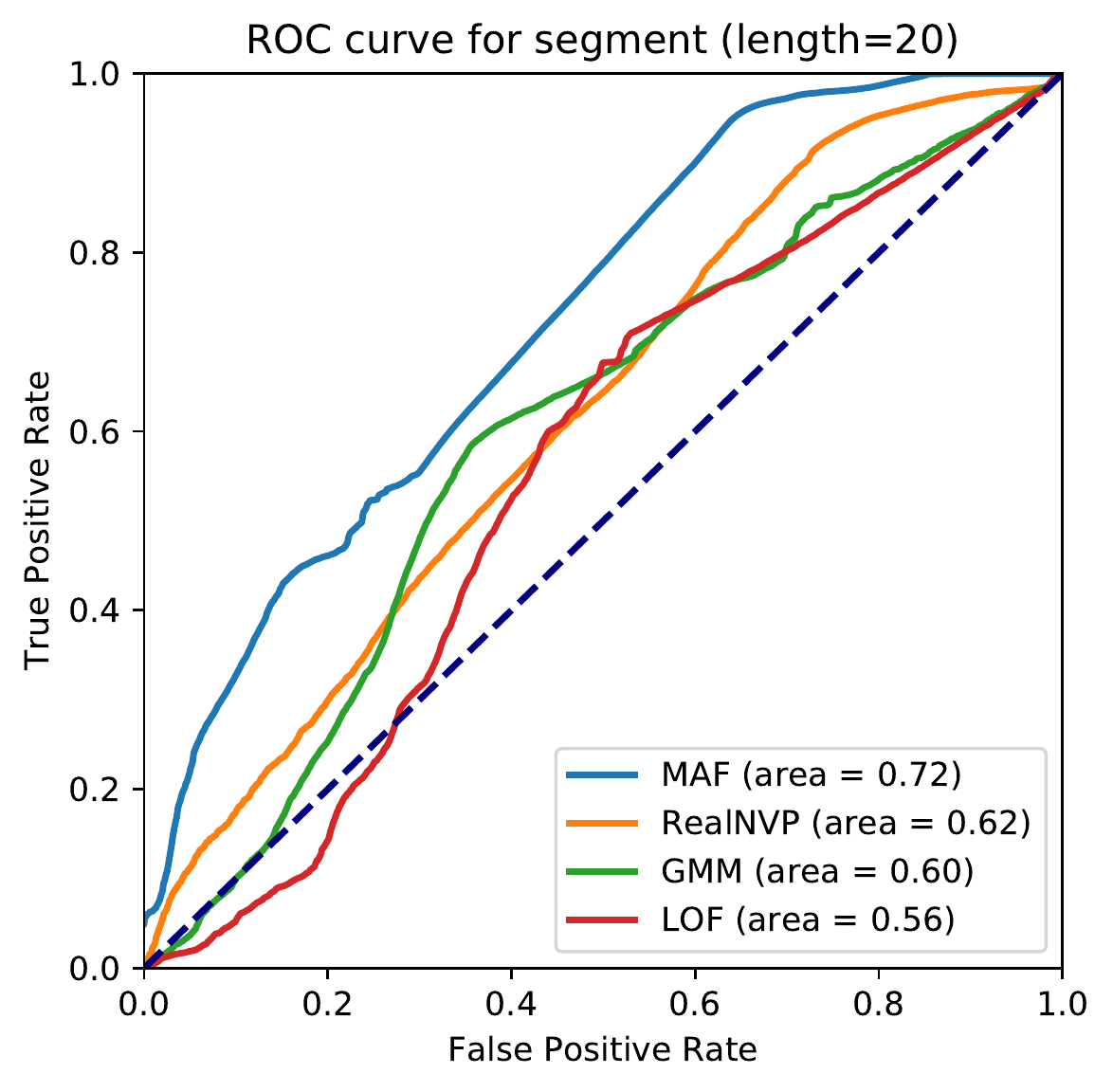}}
     \subfigure[]{\includegraphics[width=0.3\textwidth]{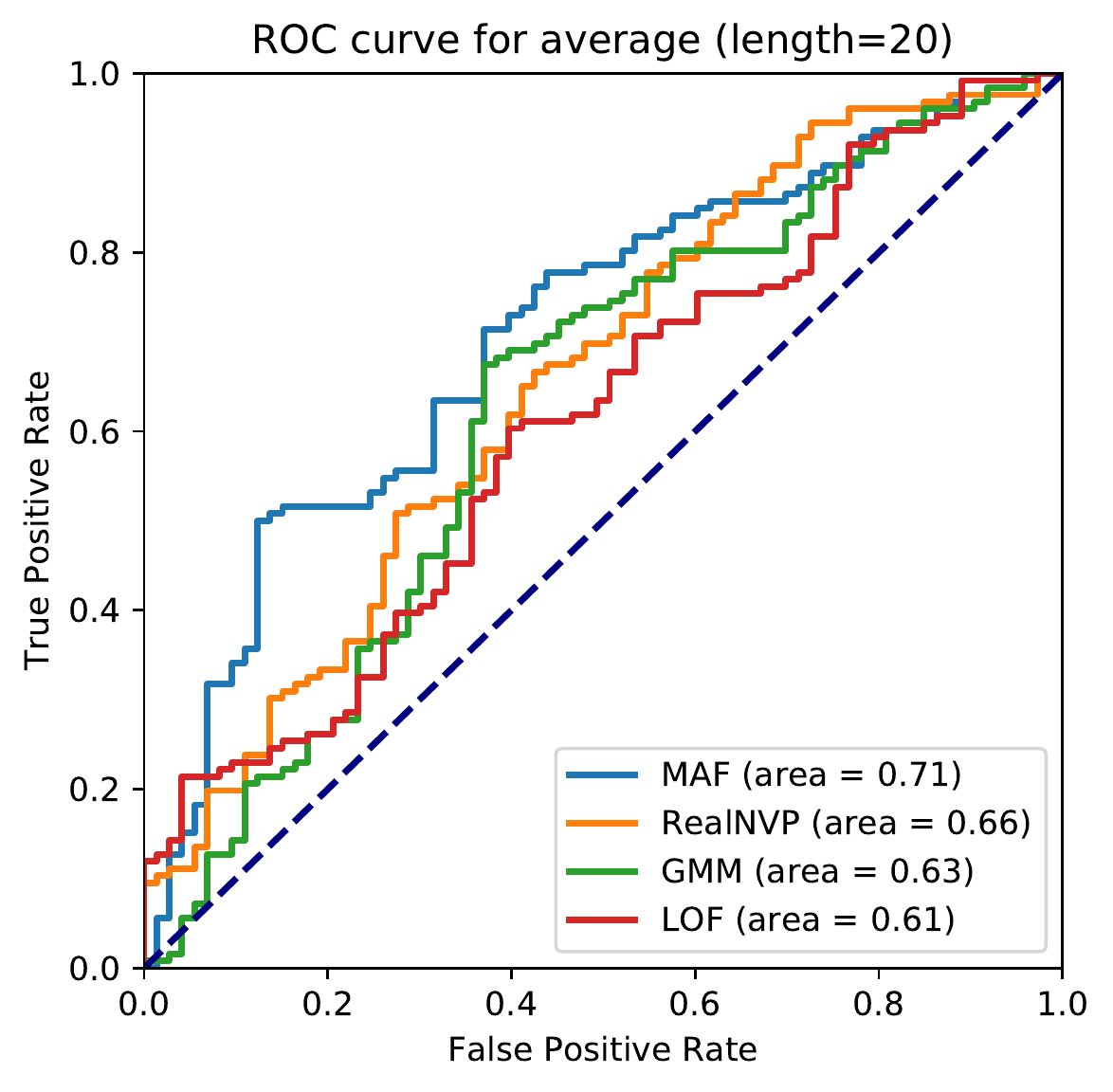}}
     \subfigure[]{\includegraphics[width=0.3\textwidth]{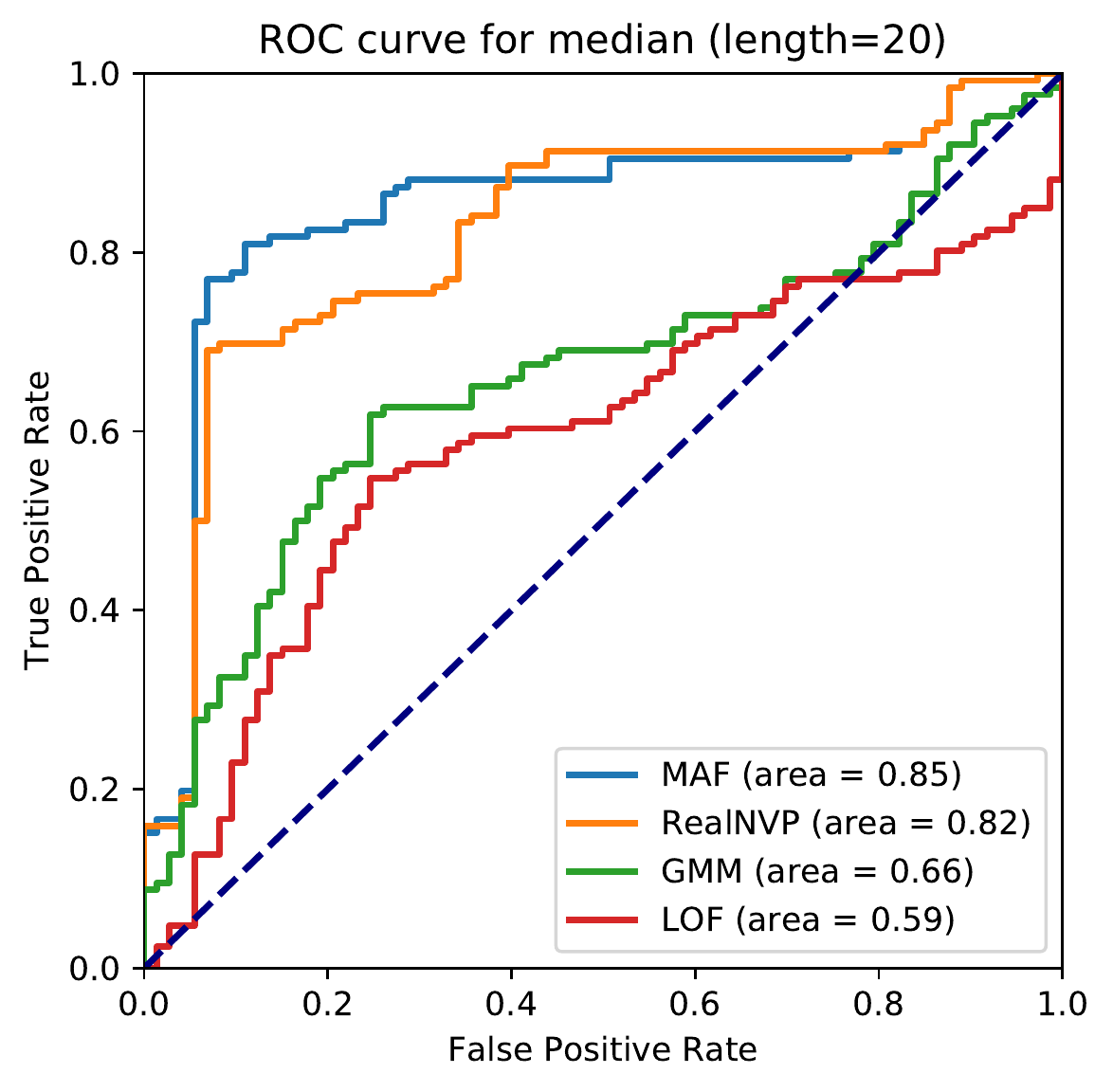}}
     
     \subfigure[]{
     \includegraphics[width=0.3\textwidth]{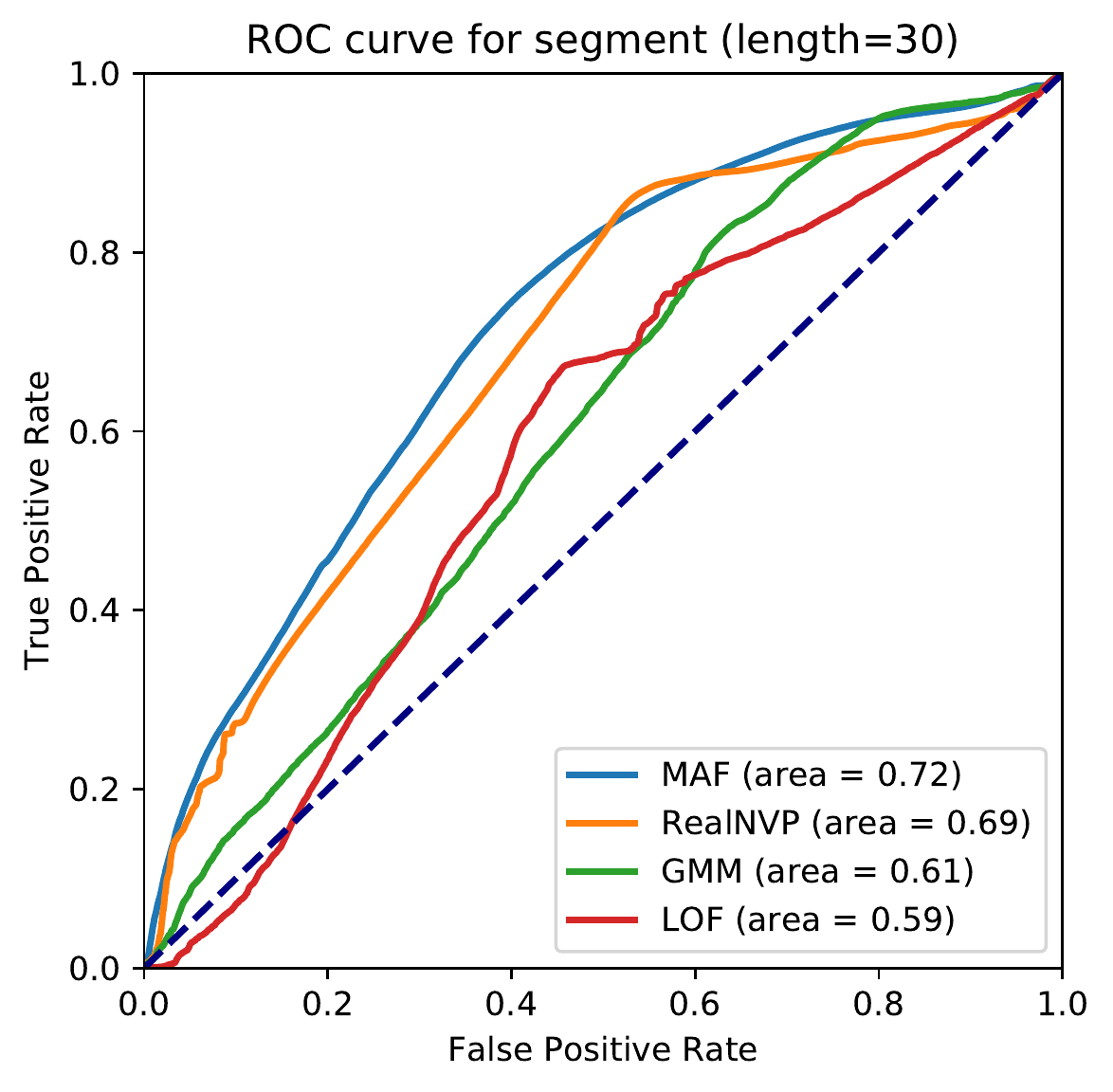}}
     \subfigure[]{\includegraphics[width=0.3\textwidth]{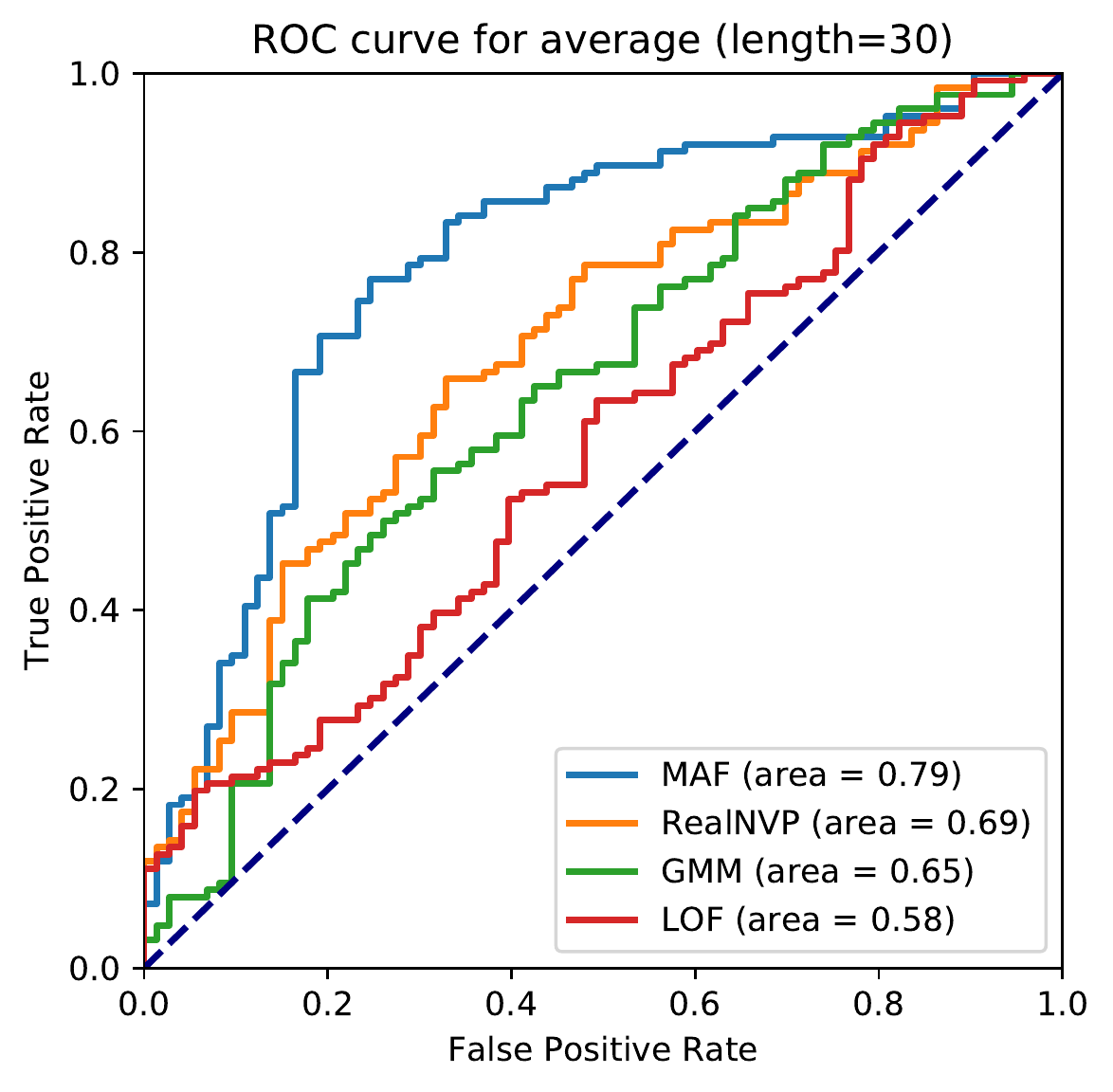}}
     \subfigure[]{\includegraphics[width=0.3\textwidth]{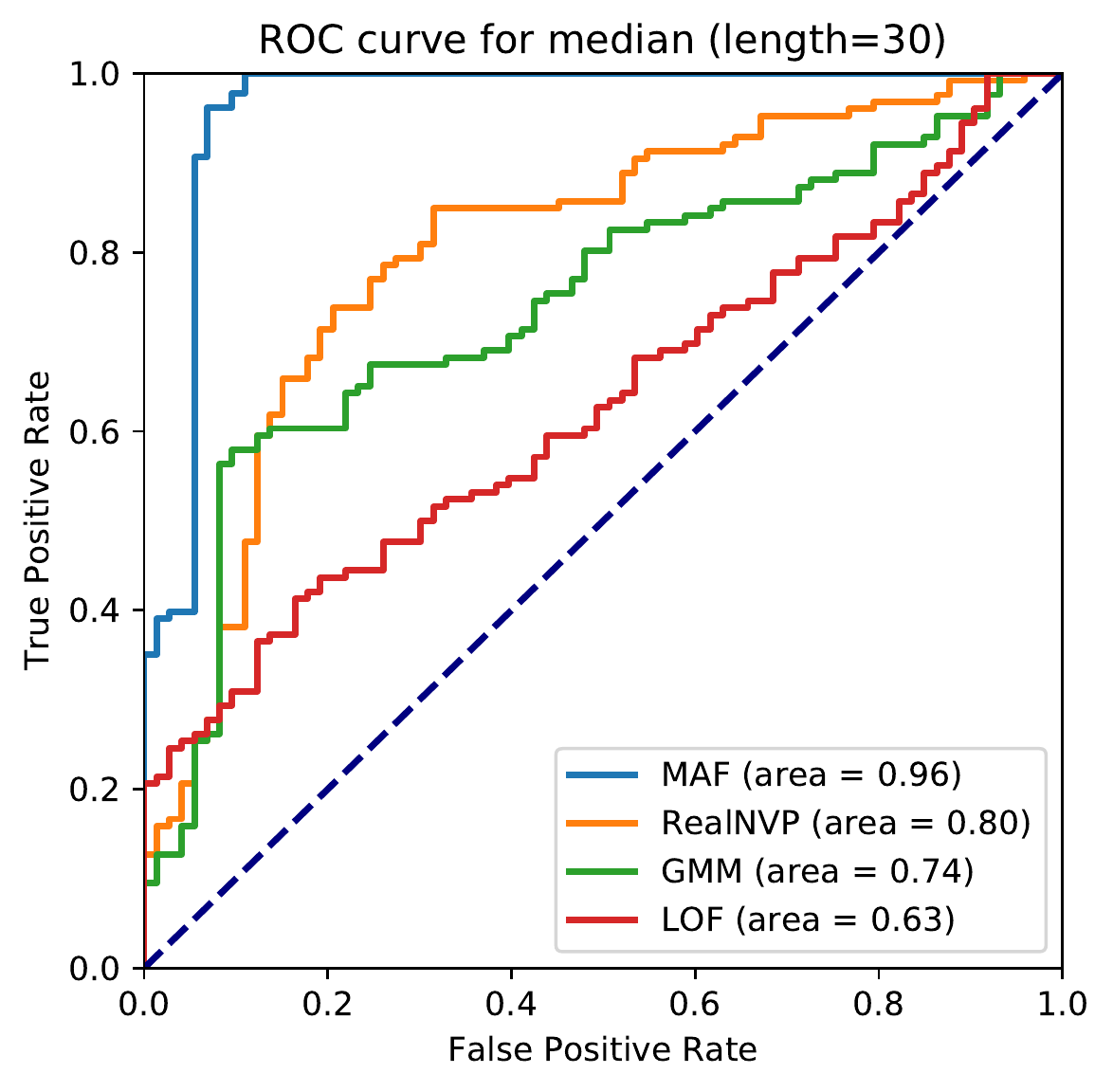}}
     \caption{Anomaly  detection  results  for  BUS $\times$ CAR scenario. ROC curves and respective AUROC values for segments (left column) and for trajectories, using average (middle column), and median (right column). The rows represent the segment lengths -- $10$ (a, b, c), $20$ (d, e, f), and $30$ (g, h, i). The dashed line indicates a completely random detector.}
    \label{fig:roc-2}
\end{figure*}

\section{Conclusion and Further Work}

Anomaly detection is a challenging task with important practical applications. In the context of trajectory data, GPS measurements are usually widely available. However, the high dimensional patterns and the lack of labeled data hinder the application of standard techniques.

In this work we have proposed \mymethodAC, an unsupervised density estimation methodology that includes flexible normalizing flows, more specifically the Real NVP and the MAF structures. \mymethodAC ~aggregates the analytical log-likelihood values of trajectory segments into a single robust anomaly score, which enables the use of trajectories with distinct lengths. The empirical results obtained using real world data showed promising performance compared to the LOF and GMM baselines, specially when considering the autoregressive MAF-based version.

The present research outcome encourages us to pursue additional NF approaches for trajectory anomaly detection. For instance, future work shall evaluate the use of convolution-based flows, such as the so-called Glow \cite{kingma2018glow}, which can handle data with multiple channel representation. Models with more complex invertible transformations, such as the recently proposed \cite{huang2018neural,oliva2018transformation,durkan2019neural}, are also worthy subjects of future investigations.

\bibliographystyle{IEEEtran}
\bibliography{references}

\end{document}